\documentclass[10pt,twocolumn,letterpaper]{article}

\usepackage[pagenumbers]{cvpr} % To force page numbers, e.g. for an arXiv version

\definecolor{cvprblue}{rgb}{0.21,0.49,0.74}
\usepackage[pagebackref,breaklinks,colorlinks,allcolors=cvprblue]{hyperref}

\def\paperID{20312} % *** Enter the Paper ID here
\def\confName{CVPR}
\def\confYear{2026}

\usepackage{graphicx}  % 核心宏包，必须包含，用于插入图片

\title{
  \includegraphics[width=1.2cm, keepaspectratio]{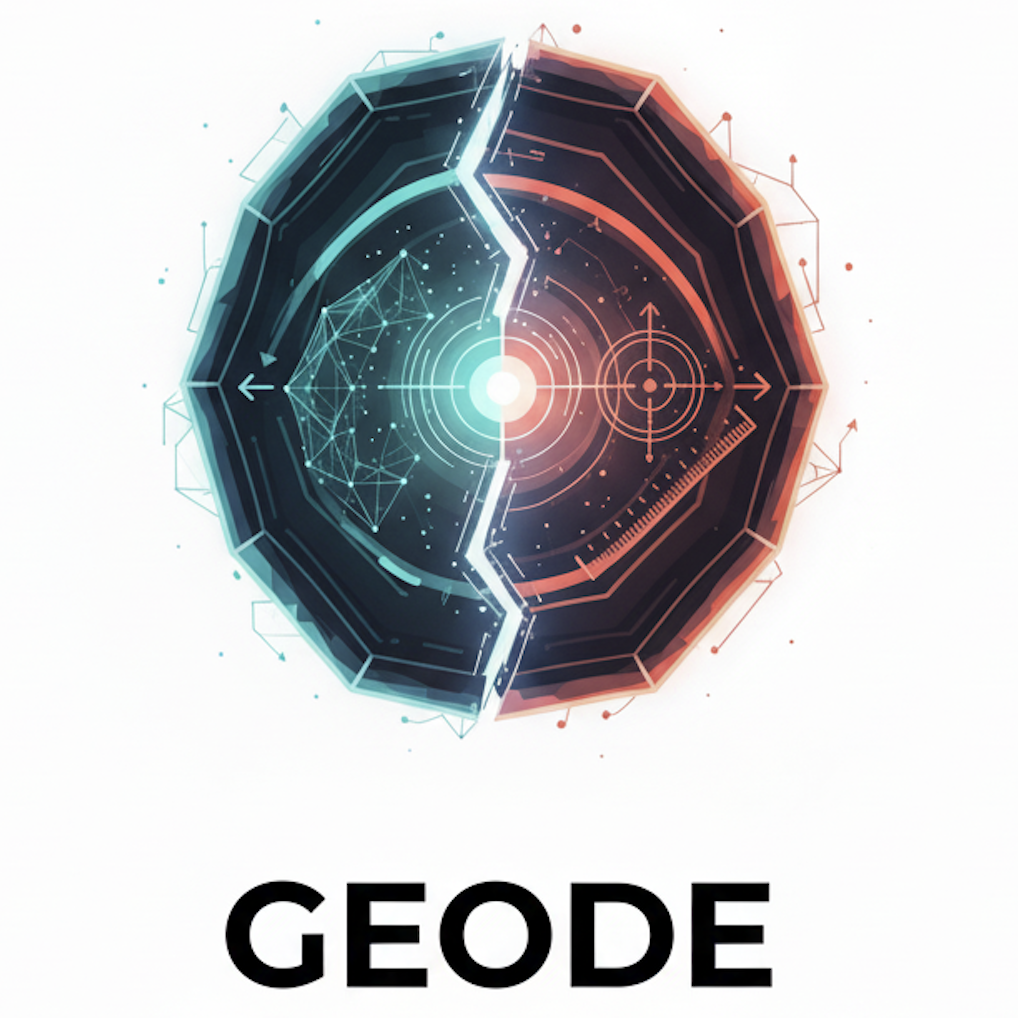} 
  Beyond Flatlands: Unlocking Spatial Intelligence by Decoupling 3D Reasoning from Numerical Regression
}

\author{Zhongbin Guo\thanks{Equal contribution.}~,~Jiahe Liu\footnotemark[1]~,~Yushan Li\footnotemark[1]~,~Wenyu Gao,~Zhen Yang,~Chenzhi Li,~Xinyue Zhang,~Ping Jian\thanks{Corresponding Author.}\\
School of Computer Science \& Technology\\
Beijing Institute of Technology\\
{\tt\small \{guozhongbin,~pjian\}@bit.edu.cn}
}

\usepackage{multirow}
\newcommand{\code}[1]{\texttt{#1}}
\usepackage{arydshln}
\begin{document}
\maketitle
\begin{abstract}
Existing Vision Language Models (VLMs) architecturally rooted in ``flatland" perception, fundamentally struggle to comprehend real-world 3D spatial intelligence. This failure stems from a dual-bottleneck: input-stage conflict between computationally exorbitant geometric-aware encoders and superficial 2D-only features, and output-stage misalignment where discrete tokenizers are structurally incapable of producing precise, continuous numerical values. To break this impasse, we introduce \textbf{GEODE} (Geometric-Output and Decoupled-Input Engine), a novel architecture that resolves this dual-bottleneck by decoupling 3D reasoning from numerical generation. GEODE augments main VLM with two specialized, plug-and-play modules: Decoupled Rationale Module (DRM) that acts as spatial co-processor, aligning explicit 3D data with 2D visual features via cross-attention and distilling spatial Chain-of-Thought (CoT) logic into injectable Rationale Tokens; and Direct Regression Head (DRH), an ``Embedding-as-Value" paradigm which routes specialized control tokens to a lightweight MLP for precise, continuous regression of scalars and 3D bounding boxes. The synergy of these modules allows our 1.5B parameter model to function as a high-level semantic dispatcher, achieving state-of-the-art spatial reasoning performance that rivals 7B+ models.
\end{abstract}    
\section{Introduction}
\label{sec:intro}

\begin{figure}[t]
    \vspace{-2.5em}
    \centering
    \includegraphics[width=\linewidth]{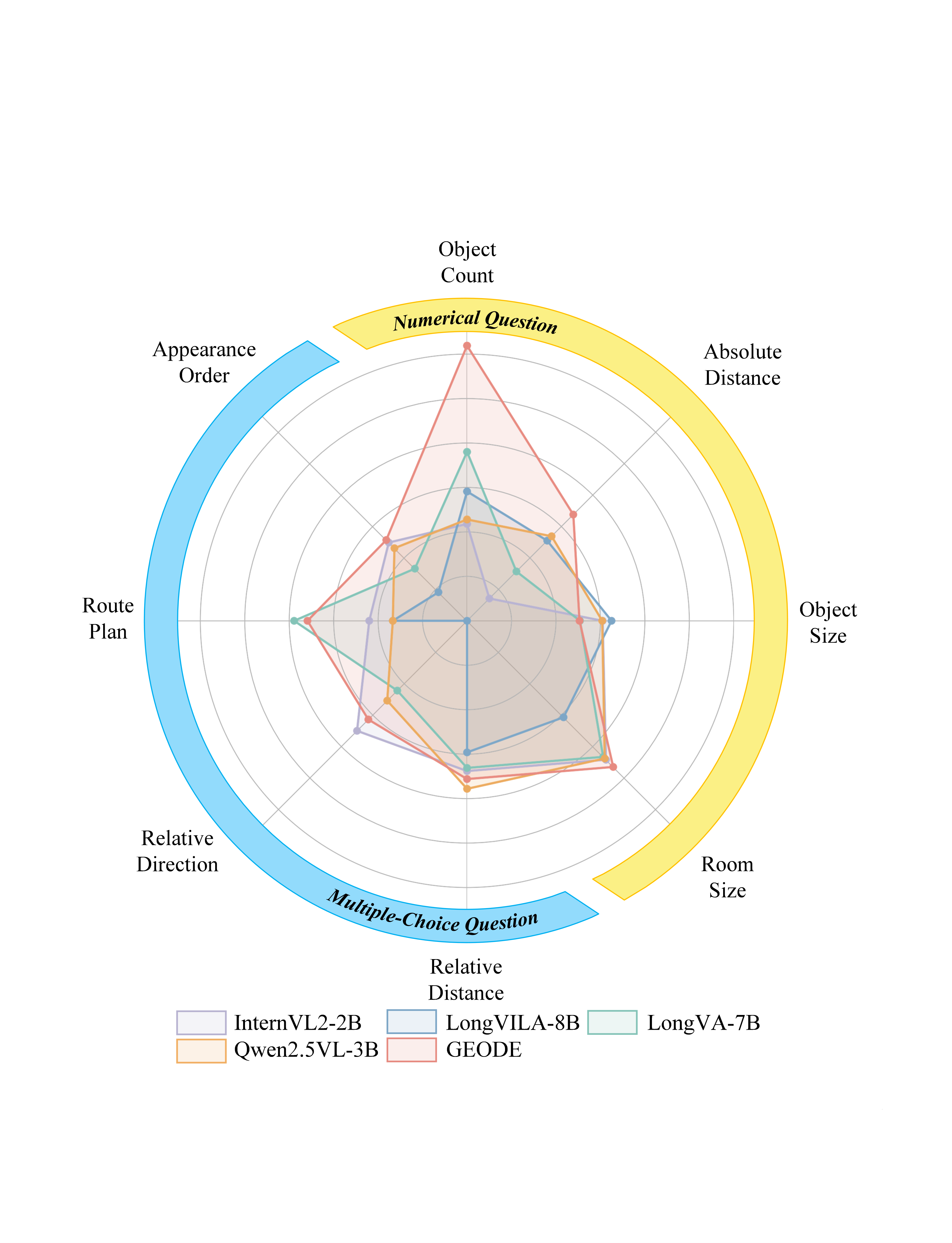}
    \vspace{-5em}
    \caption{Performance comparison of GEODE with other mainstream VLM models on VSI-bench~\cite{yangThinkingSpaceHow2024}. Our model with only 1.5B parameters achieves SOTA overall performance, especially on Object Count and Absolute Distance tasks.} 
    \label{fig1}
\end{figure}

Recent advances in Vision Language Models (VLMs)~\cite{baiQwen25VLTechnicalReport2025,openaiGPT4.5,tong2024cambrian,claude-4-5-sonnet} have marked a pivotal moment in artificial intelligence, demonstrating remarkable capabilities in understanding and reasoning over 2D images, videos, and text. However, the frontier of multimodal artificial intelligence is rapidly expanding beyond the flatland of pixels into the physically grounded, dynamic, and three-dimensional world we inhabit~\cite{zhengMultimodalSpatialReasoning2025}. This transition presents a fundamental challenge that transcends mere data scaling: current VLM architectures, architecturally rooted in 2D perception, are overwhelmingly trained on planar tasks such as OCR and 2D grounding~\cite{zhang2024vision,guo2025tamms}, yet they struggle to comprehend and reason about explicit geometric, spatial structures and temporal dynamics~\cite{liSTIBenchAreMLLMs2025}. Their reliance on patch-level visual tokens makes them susceptible to spatial hallucinations — producing interpretations that are geometrically inconsistent or physically implausible~\cite{yangThinkingSpaceHow2024,zhangFlatlandSpaceTeaching2025}. Consequently, they exhibit poor performance on real-world 3D tasks such as depth estimation, distance prediction, and path planning, severely limiting their deployment in embodied intelligence and autonomous driving scenarios~\cite{liuNavR1ReasoningNavigation2025,dangRynnECBringingMLLMs2025}.

To address this deficiency, current approaches face two fundamental bottlenecks. At the \textit{input stage}, a critical trade-off arises between geometric fidelity and computational tractability. While extending 2D-VLM with more spatio-temporal data~\cite{liu2025videoxlproreconstructivetokencompression} is computationally manageable, it remains architecturally insufficient for interpreting explicit 3D geometry. Conversely, attempts to directly integrate high-fidelity 3D sensors like LiDAR point clouds~\cite{wang2018lidar,xu2024pointllm,liu2024uni3d} achieve genuine geometric awareness, but at the cost of prohibitive computational overheads that render them impractical for scalable deployment. Concurrently, at the \textit{output stage}, a structural misalignment emerges between the discrete, token-based nature of language models and the continuous values of the physical world. The autoregressive generation of tokens is fundamentally ill-suited for producing precise numerical quantities for tasks like distance measurement or object localization~\cite{wen2025diffusionvla}. This forces an untenable compromise: models must either quantize continuous values into a coarse, error-prone vocabulary or abandon numerical precision entirely. Consequently, the prevailing design leaves VLMs trapped in a dilemma, being either geometrically superficial, computationally infeasible, or numerically imprecise.

To break this impasse and truly unlock spatial intelligence, we introduce \textbf{GEODE (Geometric-Output and Decoupled-Input Engine)}, a novel architecture that fundamentally resolves this dual-bottleneck by decoupling 3D reasoning from numerical generation. GEODE transforms the main VLM with only 1.5B parameters from a monolithic processor into a parameter-efficient orchestrator, augmenting it with two specialized, plug-and-play modules.

First, to address the \textit{input-stage} deficiency, we present \textbf{Decoupled Rationale Module (DRM)}, a decoupled spatial reasoning co-processor. Unlike approaches that rely on implicit 2D geometric cues, DRM achieves a high-fidelity 2D-3D alignment by fusing native visual features with explicit encoded 3D data, such as point clouds and camera poses, through a cross-modal attention mechanism. This module is subsequently trained on spatial Chain-of-Thought (CoT) data to distill its complex spatio-temporal reasoning logic into a set of compact, injectable Rationale Tokens. These tokens, which encapsulate the reasoning process of the 3D world, provide the main VLM with rich, pre-computed spatial context, thereby jointly solving the geometrically superficial and computationally infeasible dilemma.

Second, to achieve holistic numerical decoding for 3D spatial tasks, we propose the \textbf{Direct Regression Head (DRH)}. This ``Embedding-as-Value" paradigm fundamentally bypasses the tokenization bottleneck for continuous numerical values. Our VLM is trained to emit specialized control tokens when a quantitative answer is required. The hidden-state embedding of these tokens is then intercepted and routed away from the standard language head to the DRH — a lightweight, specialized MLP trained with a regression loss. This mechanism demonstrates powerful generalization, seamlessly decoding embeddings into both precise scalar values for distance and multi-dimensional vectors for 3D bounding boxes. This solves the final numerically imprecise problem by design.

\begin{figure*}[!t]
    \centering
    % \vspace{-0.3cm}
    \includegraphics[width=0.99\textwidth]{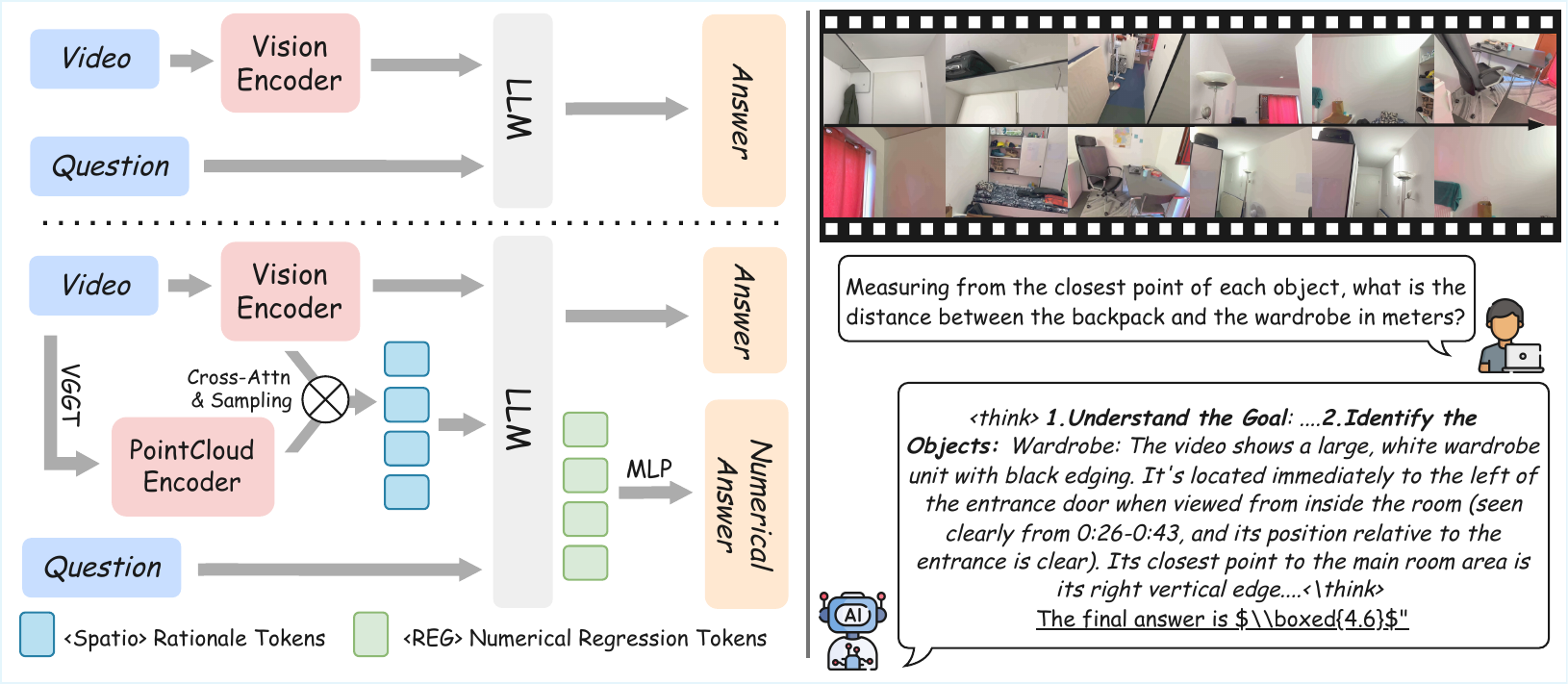}
    % \vspace{-0.5cm}
    \caption{
        Overview of the GEODE architecture, contrasted with standard VLMs.
        (\textbf{Top-Left}) Standard VLMs are architecturally rooted in 2D perception, and their discrete tokenizers are ill-suited for generating precise, continuous numerical values.
        (\textbf{Bottom-Left}) GEODE architecture resolves this dual-bottleneck by decoupling 3D reasoning from numerical generation. \textbf{Decoupled Rationale Module (DRM)} acts as spatial co-processor, fusing 2D visual features with 3D pointcloud data and distilling the logic into injectable \texttt{<Spatio>} Rationale Tokens to solve the input bottleneck. \textbf{Direct Regression Head (DRH)} implements ``Embedding-as-Value" paradigm, intercepting \texttt{<REG>} control tokens and regressing their embeddings directly into continuous values to solve the output bottleneck.
        (\textbf{Right}) A Spatial Chain-of-Thought (CoT) sample used for training DRM to generate Rationale Tokens encapsulate the underlying reasoning process.
    }
    % \caption{Overview of Video-XL. Video-XL utilizes a unified visual encoding scheme for single images, multi images, and videos. Visual tokens are first split into intervals by Dynamic chunk adjustor. Then, in LLM compressor, the activations from the visual contexts in each chunk are condensed into Visual Summarization Tokens (VSTs). Conditioned on the VSTs from previous chunks, Video-XL performs next-token prediction auto-regressively.  }
    \label{fig:overview}
    \vspace{-0.3cm}
\end{figure*}

The synergy of the DRM (input rationale) and DRH (output regression) modules creates a remarkably parameter-efficient framework. The 1.5B main LLM~\cite{yang2024qwen2} is liberated from the burdens of low-level 3D processing and high-precision numerical generation, allowing it to function as a high-level semantic dispatcher. This architecture allows our model, aligned from scratch, to achieve spatial intelligence on-par with, and in some cases exceeding, 7B+ parameter models on complex benchmarks like VSI-Bench~\cite{yangThinkingSpaceHow2024}, as shown in Fig \ref{fig1}.

\noindent Our main contributions are summarized as follows: 
\begin{itemize} 
\item We propose \textbf{GEODE}, a novel dual-decoupling architecture that resolves the fundamental input-reasoning and output-generation bottlenecks for spatial intelligence. 
\item We introduce \textbf{Decoupled Rationale Module (DRM)}, a plug-and-play co-processor aligns explicit 3D point cloud features with 2D vision via cross-attention and distills CoT-based spatial logic into injectable Rationale Tokens. 
\item We present \textbf{Direct Regression Head (DRH)}, an ``Embedding-as-Value" mechanism that enables precise, continuous regression of both scalar values and 3D bboxes by decoding specialized token embeddings. 
\item We demonstrate state-of-the-art parameter efficiency, showing our 1.5B model rivals 7B+ models on challenging 3D spatial reasoning benchmarks by integrating our decoupled modules. 
\end{itemize}

\section{Methodology}
\label{sec:method}

Our \textbf{GEODE} framework is a dual-decoupling architecture which transforms a parameter-efficient VLM into high-level spatial orchestrator. It achieves this by introducing two synergistic components: Decoupled Rationale Module (DRM) for input-stage reasoning, and Direct Regression Head (DRH) for output-stage numerical generation. This section details the design of each module.

\subsection{Decoupled Rationale Module (DRM)}

The DRM is a plug-and-play co-processor designed to resolve the input-stage bottleneck. Its goal is to perform complex spatio-temporal reasoning externally to the main VLM and distill this logic into a set of compact, injectable \textbf{Rationale Tokens}. This process involves two key phases: High-Fidelity 3D-2D Feature Fusion and Rationale-Guided Contrastive Alignment.

\subsubsection{High-Fidelity 3D-2D Feature Alignment}
To achieve genuine spatial understanding, the DRM must explicitly fuse 2D appearance features with 3D geometric and pose data, to compensate for the lack of learning of 3D data during its training process.

\noindent\textbf{3D Spatio-Temporal Context.}
We first extract explicit 3D structure from the input video frames $\{v_i\}_{i=1}^N$. We employ pretrained geometry transformers VGGT~\cite{wangVGGTVisualGeometry2025} as reconstruction engine to process the video and generate a 3D point cloud $\mathcal{P}$ along with predicted camera poses $\mathcal{C}$.
\begin{equation}
    \{\mathcal{P}, \mathcal{C}\} = f_{\text{VGGT}}(\{v_i\}_{i=1}^N)
\end{equation}

\noindent To encode this data into semantic-aware representation, we utilize Sonata~\cite{wuSonataSelfSupervisedLearning2025} as dedicated 3D encoder, explicitly processes both pointcloud geometry and camera pose information, yields $K$ high-level 3D context tokens:
\begin{equation}
    \mathbf{F}_{\text{3D}} = f_{\text{Sonata}}(\mathcal{P}, \mathcal{C}) \in \mathbb{R}^{K \times d_{\text{3D}}}
\end{equation}

\noindent\textbf{2D Visual Features.}
Concurrently, we process the video frames with InternViT~\cite{chen2024expanding} visual encoder to extract a sequence of $T$ visual tokens:
\begin{equation}
    \mathbf{F}_{\text{2D}} = f_{\text{ViT}}(\{v_i\}_{i=1}^N) \in \mathbb{R}^{T \times d_{\text{model}}}
\end{equation}

\noindent\textbf{Bridging the 2D-3D Modality Gap.}
A simple concatenation of $\mathbf{F}_{\text{2D}}$ and $\mathbf{F}_{\text{3D}}$ would create a significant feature-space gap. To enable a more interpretable and effective fusion, we align 3D context to 2D visual space using cross-attention mechanism, while ViT's native flattened features $\mathbf{F}_{\text{2D}}$ serve as the Query and the explicit 3D features $\mathbf{F}_{\text{3D}}$ serve as the Key and Value. This allows each 2D patch token to ``query" the entire 3D point cloud, enriching itself with the most relevant geometric context. The resulting fused features $\mathbf{F}_{\text{fused}} \in \mathbb{R}^{T \times d_{\text{model}}}$ are then passed through a lightweight mamba sequence model~\cite{gu2024mamba} for final temporal alignment, producing the output spatio-temporal features $\mathbf{F}_{\text{ST}}$.

\subsubsection{Rationale-Guided Reconstruction}
\label{sec:rationale_reconstruct}
The goal of DRM is not just to fuse features, but to distill high-level reasoning. We further train DRM to generate a set of $M$ specific \textbf{Rationale Tokens} (denoted as $<\text{spatio}>$) which encapsulate reasoning process from spatial CoT data.

Inspired by SSR's~\cite{liuSSREnhancingDepth2025} methodology, we utilize a dataset of spatial reasoning questions and their corresponding CoT rationales $R = \{r_1, ..., r_L\}$ during the first stage of training. We first use a pretrained frozen text encoder $h_{\text{text}}$ to map the entire textual rationale $R$ into a high-dimensional ``ground truth" semantic embedding $\mathbf{e}_R \in \mathbb{R}$. Separately, DRM processes the corresponding 3D-2D video input and outputs its sequence of $M$ Rationale Tokens, $\mathbf{E}_{\text{spatio}} = \{\mathbf{e}_1, ..., \mathbf{e}_M\}$.

\begin{figure*}[!t]
    \centering
    % \vspace{-0.3cm}
    \includegraphics[width=0.99\textwidth]{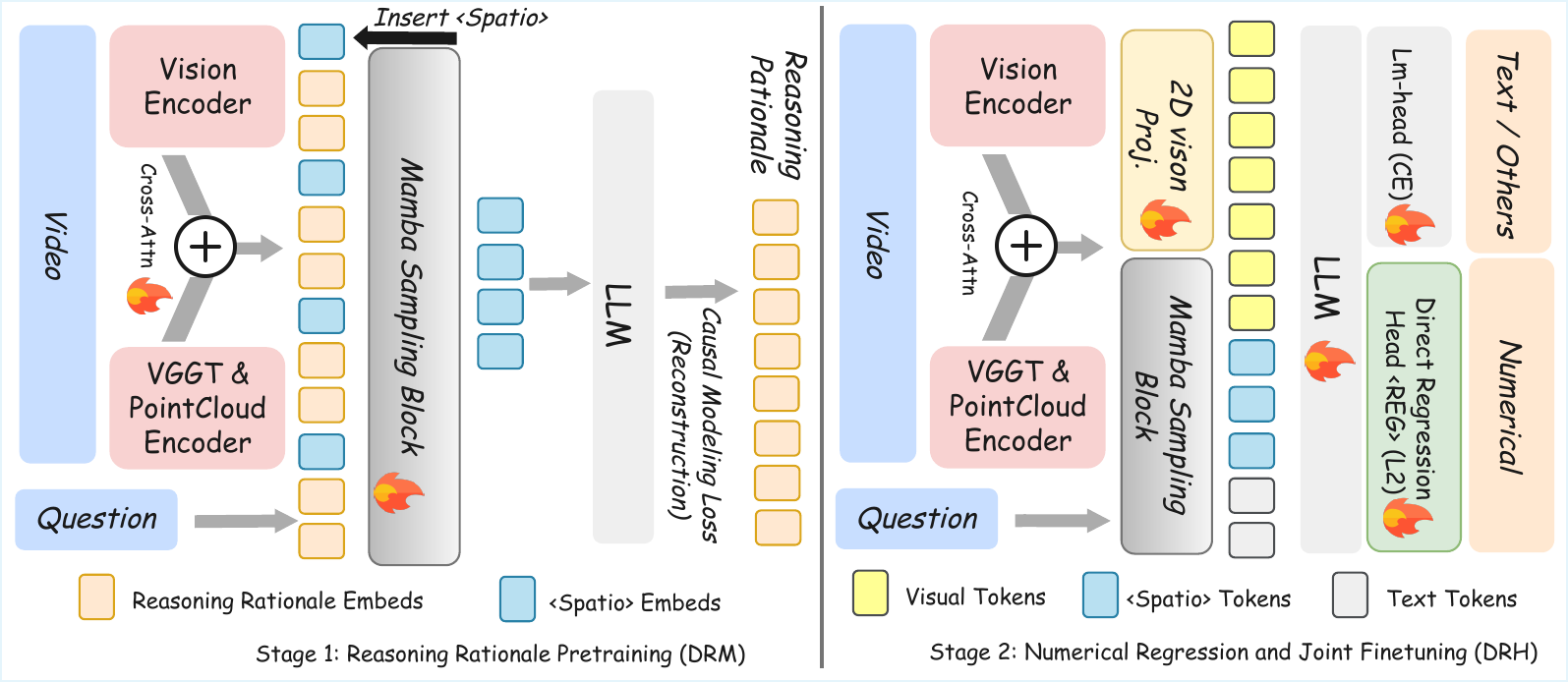}
    % \vspace{-0.5cm}
    \caption{
        The two-stage training paradigm of GEODE.
        \textbf{Stage 1: Reasoning Rationale Pretraining (DRM).}
        The main LLM parameters are frozen. Only the DRM is trained, optimized via a Rationale-Guided Reconstruction loss ($\mathcal{L}_{DRM}$) to generate \texttt{<Spatio>} embeddings that the frozen LLM can autoregressively reconstruct into the corresponding textual reasoning rationale.
        \textbf{Stage 2: Numerical Regression and Joint Finetuning (DRH).}
        The pretrained DRM is frozen, and its \texttt{<Spatio>} tokens are injected as context. The VLM backbone and the newly initialized DRH are jointly finetuned using a mixed-loss objective: Cross-Entropy ($\mathcal{L}_{CE}$) for text generation and L2 regression ($\mathcal{L}_{DRH}$) for numerical outputs routed to the DRH.
    }
    % \caption{Overview of Video-XL. Video-XL utilizes a unified visual encoding scheme for single images, multi images, and videos. Visual tokens are first split into intervals by Dynamic chunk adjustor. Then, in LLM compressor, the activations from the visual contexts in each chunk are condensed into Visual Summarization Tokens (VSTs). Conditioned on the VSTs from previous chunks, Video-XL performs next-token prediction auto-regressively.  }
    \label{fig:training_recipe}
    \vspace{-0.3cm}
\end{figure*}

These latent tokens are then fed as a prefix to the frozen $h_{\text{LLM}}$. The objective is to train the DRM to produce $\mathbf{E}_{\text{spatio}}$ such that the LLM, without any fine-tuning, can autoregressively reconstruct the original textual rationale $R$. We optimize this via a standard Causal Language Modeling loss on the rationale tokens:
\begin{equation}
    \begin{split}
    \mathcal{L}_{\text{DRM}} &= \mathcal{L}_{\text{CLM}}(h_{\text{LLM}}(\mathbf{E}_{\text{spatio}}),~R) \\
        &= - \sum_{i=1}^{L} \log p(r_i | r_{<i}, \mathbf{E}_{\text{spatio}};~\theta_{\text{DRM}})
    \end{split}
\end{equation}
where the parameters of $h_{\text{LLM}}$ are frozen and the gradient flows back to update only the parameters of DRM $\theta_{\text{DRM}}$, which forces DRM to project its spatio-temporal reasoning into a semantic space already consistent with the frozen LLM. Upon convergence, DRM is used in the second stage as a plug-and-play module, where the generated $\mathbf{E}_{\text{spatio}}$ are injected as extra context to the main LLM to facilitate grounded spatial reasoning.

\subsection{Direct Regression Head (DRH)}
\label{sec:drh}

The second bottleneck we address is the output-stage misalignment, where the discrete, autoregressive nature of LLMs is fundamentally ill-suited for generating continuous numerical values. Forcing a VLM to output a distance like ``6.5" breaks it into multiple, disconnected tokens (`6', `.', `5'), destroying the value's holistic integrity and making precise regression via cross entropy intractable.

To achieve holistic numerical decoding, we introduce the \textbf{Direct Regression Head (DRH)}, an "Embedding-as-Value" paradigm. Instead of treating numerical outputs as text, we treat them as direct regression targets. We expand the VLM's vocabulary with a set of specialized control tokens, $\langle\text{REG}\rangle$ for scalar values and $\langle\text{3DBBOX}\rangle$ for structured 3D coordinates.

During the second-stage fine-tuning, the VLM is trained on quantitative spatial data (e.g., ``What is the distance between...?" or ``Locate the table") to emit the appropriate control token as the answer. When such a token is generated, we intercept its corresponding hidden-state embedding $\mathbf{h}_{\text{control}} \in \mathbb{R}^{d_{\text{model}}}$. This embedding is routed away from the standard language modeling head and is instead passed to the DRH, $f_{\text{DRH}}$, a lightweight, task-specific MLP:
\begin{equation}
    \mathbf{y}_{\text{reg}} = f_{\text{DRH}}(\mathbf{h}_{\text{control}})
\end{equation}
where $\mathbf{y}_{\text{reg}}$ is the final continuous output vector (e.g., single scalar for distance, or 7-DoF vector for 3D bounding box).

This hybrid-output architecture is trained using a mixed-loss objective. For standard text generation, we use the Cross-Entropy loss. For quantitative tasks routed to the DRH, we apply L2 regression loss:
\begin{equation}
    \mathcal{L}_{\text{DRH}} = \mathcal{L}_{\text{L2}}(\mathbf{y}_{\text{reg}}, \mathbf{y}_{\text{gt}})
\end{equation}
This design completely bypasses the tokenization bottleneck, ensuring numerical values are generated as a single, coherent unit and allowing the model to be optimized with loss function appropriate for continuous values. This mechanism demonstrates powerful generalization, seamlessly handling both single-value scalars and multi-dimensional coordinate vectors within the same unified framework.

\subsection{Training Paradigm}
\label{sec:training}

The overall training of our \textbf{GEODE} architecture is conducted in two distinct and sequential stages, aligning with the decoupled nature of our modules.

\noindent\textbf{Stage 1: Reasoning Rationale Module Pretraining.}
The first stage is dedicated to training the \textbf{DRM}. As detailed in Section~\ref{sec:rationale_reconstruct}, we freeze the parameters of the main LLM ($h_{\text{LLM}}$) and optimize only the DRM parameters ($\theta_{\text{DRM}}$). This is achieved using the Rationale-Guided Reconstruction loss, $\mathcal{L}_{\text{DRM}}$, which compels the DRM to generate Rationale Tokens ($\mathbf{E}_{\text{spatio}}$) that can be autoregressively reconstructed into the full textual rationale by the frozen LLM. This stage effectively distills complex spatio-temporal reasoning into a set of "LLM-comprehensible" embeddings, resulting in a plug-and-play spatial co-processor.

\noindent\textbf{Stage 2: Numerical Regression and VLM Joint Fine-tuning.}
In the second stage, we freeze the parameters of the pretrained DRM, then conduct a full finetuning of the 2D vision projector, LLM backbone and the newly initialized \textbf{DRH}. VLM is trained on a mixed dataset of spatial CoT tasks (which now leverage the injected $\mathbf{E}_{\text{spatio}}$ as context) and quantitative numerical tasks. This stage is optimized using the mixed-loss objective described in Section~\ref{sec:drh}, which combines the Cross-Entropy loss ($\mathcal{L}_{\text{CE}}$) for text generation with the regression loss ($\mathcal{L}_{\text{DRH}}$) for numerical outputs. The total loss is:
\begin{equation}
    \mathcal{L}_{\text{total}} = \mathcal{L}_{\text{CE}} + \lambda \mathcal{L}_{\text{DRH}}
\end{equation}
where $\lambda$ is a hyperparameter to balance the text generation and numerical regression tasks. This paradigm trains the VLM to function as a high-level orchestrator, learning both to utilize the spatial context from the DRM and to route quantitative queries to the DRH.
\section{Experiments}
\label{sec:experiment}

\subsection{Dataset Collection}
\label{sec:datasets}

Our training paradigm is divided into distinct stages, each utilizing a curated set of datasets to train the specific components of our \textbf{GEODE} architecture.

\noindent \textbf{General Vision-Language Alignment.}
To establish the foundational alignment between the 2D visual encoder and the main LLM, we employ a diverse corpus of over 650k image and video samples. This includes approximately 281k spatial video samples from the base subset of ViCA-322K~\cite{fengVisuospatialCognitiveAssistant2025}, 120k general video caption samples from 25\% subset of Smit~\cite{Monfort_2021_CVPR} dataset, and 250k image caption samples from a 10\% subset of the Cambrian-Alignment~\cite{tong2024cambrian} dataset. This broad dataset ensures the model learns robust, general-purpose vision-language correspondences.

\noindent \textbf{DRM Rationale Pre-training.}
For the first-stage training of our \textbf{DRM}, we utilize datasets rich in spatio-temporal logic. This includes complex subset of ViCA-322K~\cite{fengVisuospatialCognitiveAssistant2025} (approx. 40k samples) which provides detailed descriptions of spatio-temporal tasks, and ViCA-Thinking dataset (approx. 2.68k samples), which contains high-quality spatial CoT rationales. This curated data is essential for training the DRM to distill complex reasoning into its injectable Rationale Tokens.

\noindent \textbf{DRH and Joint VLM Fine-tuning.}
The second-stage, which jointly trains the main VLM to use the frozen DRM and optimizes the \textbf{DRH}, leverages large-scale spatial reasoning and numerical datasets. We use the full VSI-590k dataset~\cite{yangCambrianSSpatialSupersensing2025} (approx. 590k samples) to train the DRH for its quantitative ``Embedding-as-Value" tasks (i.e., distance and 3D BBox regression). This is supplemented by the VLM-3R-Data~\cite{fanVLM3RVisionLanguageModels2025} (approx. 330k samples) which provides a rich set of spatial understanding, multiple-choice, and inference tasks that teach the VLM to effectively utilize the DRM's contextual Rationale Tokens.

\subsection{Implementation Details}
\label{sec:imp_details}

Our \textbf{GEODE} framework is built by aligning a vision encoder with an LLM backbone, starting from pre-trained weights. We utilize Qwen2.5-1.5B-Instruct~\cite{yang2024qwen2} as our main LLM, InternViT-300M-448px-V2\_5~\cite{chen2024expanding} as 2D vision encoder. For DRM, we employ the pretrained VGGT-1B~\cite{wangVGGTVisualGeometry2025} model to reconstruct 3D pointclouds and predict camera poses from input videos, and Sonata~\cite{wuSonataSelfSupervisedLearning2025} model as our dedicated point cloud encoder. All experiments are conducted on a single node equipped with 8$\times$A100 (80GB) GPUs. Further details regarding hyperparameters, optimizer settings, and all training stages are available in the Appendix.

\subsection{Main Results}
We evaluate GEODE’s spatial reasoning capability on VSI-Bench~\cite{yangThinkingSpaceHow2024}, whichcomprises over 5,000 QA pairs from egocentric videos in ScanNet~\cite{dai2017scannet}, ScanNet++~\cite{yeshwanth2023scannet++}, and ARKitScenes~\cite{baruch2021arkitscenes}. VSI-Bench provides two answer formats, multiple choice (MCA) and numerical (NA). Results in Tab~\ref{tab:main_results} demonstrate that GEODE achieves impressive performance, particularly on the Absolute Distance and Relation Direction tasks, which require complex spatial reasoning and precise 3D object localization. Moreover, with a parameter scale of 1.5B, GEODE achieves an overall score of 40 — performance that even outperforms some 7B models and larger closed-source counterparts.

\begin{table*}[t]
\centering
\small
\setlength{\tabcolsep}{2.5pt}
\caption{\textbf{Evaluation on VSI-Bench~\cite{yangThinkingSpaceHow2024}. GEODE achieves state-of-the-art performance on overall tasks.}}

\begin{tabular}{lcc| cccc cccc}
\toprule

% ---------- 三列同时合并两行 ----------
\multirow{2}{*}{\centering\large\textbf{Model}}   % ← 字体更大
& \multirow{2}{*}{\centering\textbf{Rank}}
& \multirow{2}{*}{\centering\textbf{Overall}}
& \multicolumn{4}{c}{\textbf{Numerical Question}}
& \multicolumn{4}{c}{\textbf{Multiple-Choice Question}} \\
\cmidrule(lr){4-7} \cmidrule(lr){8-11}

& & &
\textbf{Obj. Cnt} &
\textbf{Abs. Dist} &
\textbf{Obj. Size} &
\textbf{Roo. Size} &
\textbf{Rel. Dist} &
\textbf{Rel. Dir} &
\textbf{Rou. Plan} &
\textbf{App. Order} \\
\midrule

% GPT-4o
% & 2 & 34.0
% & 46.2 & 5.3 & \textbf{43.8} & \textbf{38.2}
% & 37.0 & 41.3 & 31.50 & 28.5 \\

% Qwen2.5VL-3B
% & 5 & 28.3
% & 22.8 & 23.1 & 16.7 & 25.4
% & 37.8 & 43.8 & 30.4 & 26.9 \\

% Qwen2.5VL7B
% & 3 & 33.0
% & 40.9 & 14.8 & 43.4 & 10.7
% & \textbf{38.6} & 38.5 & \textbf{33.0} & 29.8 \\

% \hdashline

% SFT only
% & 4 & 31.2
% & \textbf{61.8} & 11.3 & 35.8 & 11.4
% & 26.5 & 45.8 & 29.4 & 27.5 \\

% \textbf{GEODE~(1.5B)}
% & 1 & \textbf{37.0}
% & 61.7 & \textbf{25.7} & 35.9 & 31.4
% & 35.6 & \textbf{46.5} & 25.3 & \textbf{33.8} \\

% --- Baseline Section from Image ---
\multicolumn{11}{l}{\textit{Baseline}} \\
Chance Level (Frequency)
& - & 34.0
& 62.1 & 32.0 & 29.9 & 33.1
& 25.1 & 47.9 & 28.4 & 25.2 \\
\midrule

% --- Proprietary Models Section ---
\multicolumn{11}{l}{\textit{Proprietary Models (API)}} \\
GPT-4o
& 5 & 34.0  % Rank updated from 2 to 3 based on image
& 46.2 & 5.3 & 43.8 & 38.2
& 37.0 & 41.3 & 31.5 & 28.5 \\
\midrule

% --- Open-sourced VLMs Section ---
\multicolumn{11}{l}{\textit{Open-sourced VLMs}} \\

InternVL2-2B~\cite{chen2024expanding}
& 12 & 27.4 & 21.8
& 24.9 & 22.0 & 35.0 & 33.8
& 44.2 & 30.5 & 7.1 \\

InternVL2-8B~\cite{chen2024expanding}
& 4 & 34.6
& 23.1 & 28.7 & 48.2 & \textbf{39.8}
& 36.7 & 30.7 & 29.9 & \textbf{39.6} \\

InternVL2-40B~\cite{chen2024expanding}
& 2 & 36.0
& 34.9 & 26.9 & 46.5 & 31.8
& 42.1 & 32.2 & 34.0 & \textbf{39.6} \\

LongVILA-8B~\cite{chen2024longvila}
& 10 & 21.6
& 29.1 & 9.1 & 16.7 & 0.0
& 29.6 & 30.7 & 32.5 & 25.5 \\

VILA-1.5-40B~\cite{lin2024vila}
& 13 & 31.2
& 22.4 & 24.8 & \textbf{48.7} & 22.7
& 40.5 & 25.7 & 31.5 & 32.9 \\

LongVA-7B~\cite{zhang2024long}
& 8 & 29.2
& 38.0 & 16.6 & 38.9 & 22.2
& 33.1 & 43.3 & 25.4 & 15.7 \\

LLaVA-NeXT-Video-7B
& 3 & 35.6
& 48.5 & 14.0 & 47.8 & 24.2
& \textbf{43.5} & 42.4 & \textbf{34.0} & 30.6 \\

LLaVA-OneVision-7B~\cite{li2024llava}
& 7 & 32.4
& 47.7 & 20.2 & 47.4 & 12.3
& 42.5 & 35.2 & 29.4 & 24.4 \\

Qwen2.5VL-3B~\cite{baiQwen25VLTechnicalReport2025}
& 11 & 28.3
& 22.8 & 23.1 & 16.7 & 25.4
& 37.8 & 43.8 & 30.4 & 26.9 \\

Qwen2.5VL7B~\cite{baiQwen25VLTechnicalReport2025}
& 6 & 33.0  % Rank updated from 3 to 5 based on image
& 40.9 & 14.8 & 43.4 & 10.7
& 38.6 & 38.5 & 33.0 & 29.8 \\

\hdashline

% --- User's Custom/SFT Section ---
\multicolumn{11}{l}{\textit{SFT only / GEODE}} \\
SFT only
& 8 & 31.2
& 61.7 & 11.3 & 35.8 & 11.4
& 26.5 & 45.8 & 29.4 & 27.5 \\

\textbf{GEODE~(1.5B)}
& \textbf{1} & \textbf{37.0}
& \textbf{61.9} & \textbf{25.7} & 35.9 & 31.4
& 35.6 & \textbf{46.5} & 25.3 & 33.8 \\

\bottomrule
\end{tabular}
\label{tab:main_results}
\end{table*}

% \url{https://www.computer.org/about/contact}.
\section{Analysis and Ablation Studies}
\label{sec:analysis}

Our main results on VSI-Bench \cite{yangThinkingSpaceHow2024}, presented in Table \ref{tab:main_results}, demonstrate that GEODE achieves state-of-the-art performance, rivaling 7B+ models with only 1.5B parameters. This section dissects the source of this parameter-efficient performance by validating the core tenets of our "dual-decoupling" architecture. 

We conduct a series of targeted ablation studies to answer three critical questions:
(1) Is our input-stage 3D-2D fusion (DRM) computationally efficient and robust to sparse inputs?
(2) Do the \code{<Spatio>} Rationale Tokens, distilled by the DRM, effectively enhance the LLM's spatial reasoning capabilities?
(3) Are both the DRM (input decoupling) and DRH (output decoupling) necessary, and do they exhibit a synergistic effect that validates our overall design?

To ensure computational tractability, all ablation experiments in this section are trained on a 10\% subset of the full training data, maintaining the same data distribution used for the main GEODE model.

\subsection{Input Efficiency of 3D-2D Fusion}
\label{ssec:ablation_frames}

A primary challenge in resolving the input-stage bottleneck is balancing geometric fidelity against computational feasibility. Our DRM is designed for efficiency, leveraging VGGT \cite{wangVGGTVisualGeometry2025} to reconstruct point clouds from video frames, which are then processed by the Sonata \cite{wuSonataSelfSupervisedLearning2025} encoder. 

To validate this design's input efficiency, we ablate the number of video frames (8 vs. 16) fed into VGGT/Sonata branch during DRM training. As shown in Table \ref{tab:ablation_frames}, the performance difference between the two settings is negligible. This finding is significant: it suggests our approach is robust to sparse inputs and does not rely on dense multi-view information. This is likely attributable to the strong reconstruction capabilities of VGGT, which can recover high-quality 3D geometry from as few as 8 frames, providing a sufficient 3D context for the Sonata encoder. This result powerfully validates the computational advantage of our method, proving GEODE can achieve robust spatial understanding without exorbitant input data requirements.

% Please fill in your data in the table below
\begin{table}[t]
\centering
\caption{Effect of input frame count for the 3D branch (VGGT/Sonata) on VSI-Bench (10\% data). Performance remains stable, demonstrating input efficiency.}
\label{tab:ablation_frames}
\resizebox{\columnwidth}{!}{%
\begin{tabular}{@{}l|ccc@{}}
\toprule
Setting (DRM Input) & Overall (Acc.) & Abs. Dist. & Obj. Cnt. \\
\midrule
GEODE (8 Frames)  & 32.8 & 26.4 & 53.1 \\
GEODE (16 Frames) & 33.0 & 28.0 & 53.2 \\
\bottomrule
\end{tabular}%
}
% \vspace{-0.3cm}

\end{table}

\subsection{Efficacy of \code{<Spatio>} Rationale Tokens}
\label{ssec:ablation_tokens}

The core hypothesis of the DRM is that complex spatio-temporal logic can be distilled into a compact set of injectable \code{<Spatio>} Rationale Tokens. To test the efficacy and impact of these tokens, we conduct an ablation on the number of generated tokens ($M$) during inference.

As detailed in Table \ref{tab:ablation_tokens}, increasing the number of \code{<Spatio>} tokens yields a clear performance improvement. Critically, we observe that the performance gains are more pronounced on Multiple-Choice (MCA) tasks (e.g., Route Plan, Rel. Direction) than on Numerical Answer (NA) tasks (e.g., Abs. Distance). 

This finding precisely aligns with our design rationale. The \code{<Spatio>} tokens are trained via Rationale-Guided Reconstruction (Stage 1) to encapsulate the \textit{logical process} of spatial reasoning from CoT data. Consequently, they provide the most significant boost to complex, logic-based MCA questions. Numerical tasks, in contrast, are primarily dependent on the DRH for precise regression. This differentiated improvement strongly suggests that the \code{<Spatio>} tokens are not just redundant features but are effectively providing high-level, pre-computed spatial logic that enhances the main LLM's reasoning faculties.

% Please fill in your data in the table below
\begin{table}[t]
\centering
\caption{Effect of the number ($M$) of \code{<Spatio>} Rationale Tokens on VSI-Bench (10\% data). More tokens boost reasoning, especially on MCA tasks.}
\label{tab:ablation_tokens}
\resizebox{\columnwidth}{!}{%
\begin{tabular}{@{}l|ccc@{}}
\toprule
Setting (Inference) & Overall (Acc.) & MCA (Avg. Acc.) & NA (Avg. Acc.) \\
\midrule
GEODE ($M$=4)  & 28.9 & 30.4 & 27.4 \\
GEODE ($M$=8)  & 30.2 & 31.1 & 29.3 \\
GEODE ($M$=16) & 33.0 & 34.7 & 31.3 \\
\bottomrule
\end{tabular}%
}
\end{table}

\subsection{Synergy of Decoupled Modules}
\label{ssec:ablation_modules}

Our final and most critical ablation investigates the necessity and synergy of the DRM (input decoupling) and DRH (output decoupling) modules. We compare four variants, as outlined in Table \ref{tab:ablation_modules}:
\begin{itemize}
    \item \textbf{SFT only}: The baseline model, trained with standard Supervised Finetuning (SFT) without our modules. It suffers from both the input (superficial 2D features) and output (discrete tokenizer) bottlenecks.
    \item \textbf{SFT + DRH}: This variant adds only the DRH, resolving the output bottleneck via ``Embedding-as-Value" regression but remaining a ``flatland" perception model at its input.
    \item \textbf{SFT + DRM}: This variant adds only the DRM, gaining rich 3D spatial context via \code{<Spatio>} tokens but still relying on the flawed autoregressive process to generate numerical text.
    \item \textbf{GEODE (Full)}: Our complete model, integrating both DRM and DRH to resolve the dual-bottleneck.
\end{itemize}

% Please fill in your data in the table below
\begin{table}[t]
\centering
\caption{Component ablation of DRM and DRH on VSI-Bench (10\% data). The results demonstrate a strong synergy, as the full GEODE model significantly outperforms all partial configurations.}
\label{tab:ablation_modules}
\resizebox{\columnwidth}{!}{%
\begin{tabular}{@{}l|cc|ccc@{}}
\toprule
Setting & DRM & DRH & Overall & Abs. Dist. & Route Plan \\
\midrule
1. SFT only & & & 24.8 & 14.5 & 24.7 \\
2. SFT + DRH & & \checkmark &27.4 & 27.5 & 25.1 \\
3. SFT + DRM & \checkmark & & 28.3 & 27.9 & 27.0 \\
% GEODE ($M$=16) & 33.0 & 34.7 & 31.3 \\
4. GEODE (Full) & \checkmark & \checkmark & \textbf{33.0} & \textbf{28.0} & \textbf{27.3} \\
\bottomrule
\end{tabular}%
}
\end{table}

The results in Table \ref{tab:ablation_modules} provide a clear narrative.
(1) The \textbf{SFT only} baseline performs the poorest, confirming that standard VLMs are fundamentally ill-equipped for complex 3D spatial tasks.
(2) Adding the \textbf{DRH only} provides a significant boost, particularly on numerical tasks like \textit{Abs. Dist.}, proving the superiority of our regression head over discrete tokenization for continuous values.
(3) \textbf{DRM only} performs \textbf{better overall than DRH only}. This is a key finding, suggesting that \textbf{solving the input-stage reasoning bottleneck is more critical than solving the output-stage numerical bottleneck}. A model that cannot perceive 3D space correctly cannot output a correct distance, regardless of its output mechanism.
(4) \textbf{GEODE (Full)} achieves the best performance by a significant margin, excelling on \textit{both} reasoning-heavy tasks (\textit{Route Plan}) and numerically-precise tasks (\textit{Abs. Dist.}). This demonstrates a powerful \textbf{synergy} between the two modules. The DRM provides high-quality spatial reasoning, and the DRH provides a high-precision outlet for that reasoning. 

In summary, our ablation studies validate that the ``dual-decoupling" design of GEODE is efficient, effective, and highly synergistic, providing a robust and parameter-efficient solution to unlock 3D spatial intelligence.

\section{Related work}
\label{sec:rw}

\begin{figure*}[!t]
    \centering
    % \vspace{-0.3cm}
    \includegraphics[width=0.99\textwidth]{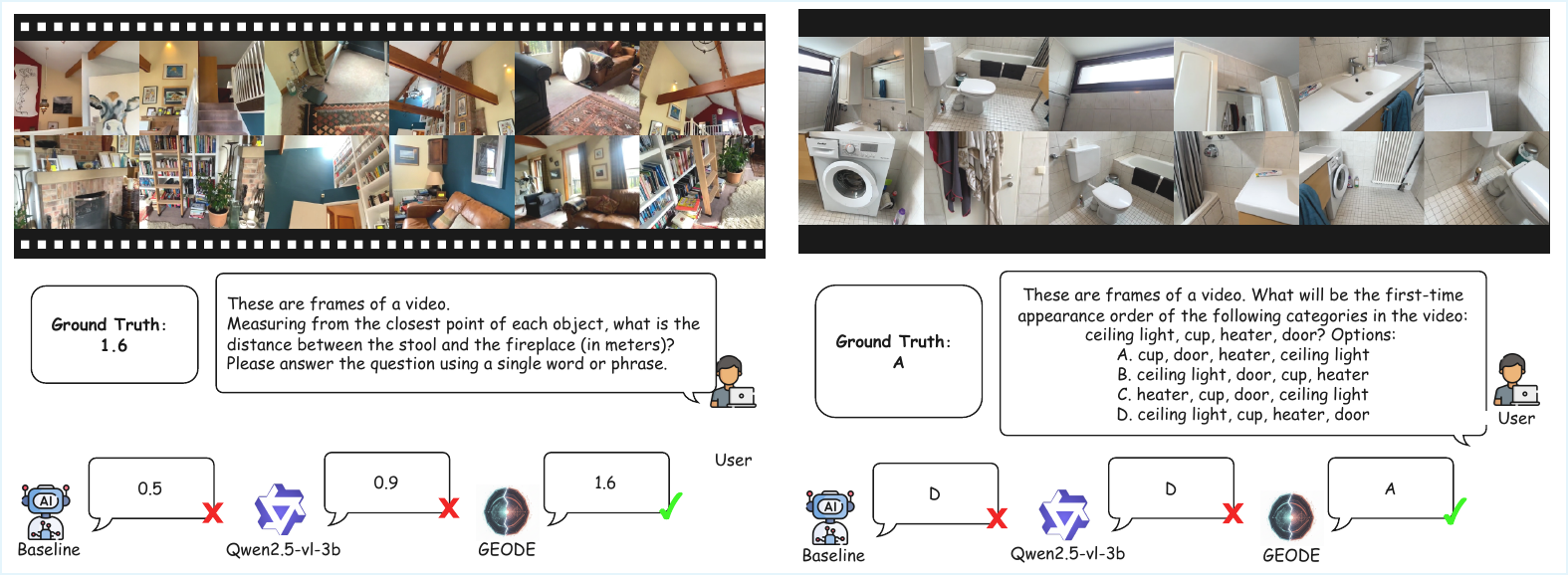}
    % \vspace{-0.5cm}
    \caption{
        Quantitive results for GEODE, SFT only baseline and Qwen2.5-VL-3B~\cite{baiQwen25VLTechnicalReport2025}. 
    }
    \label{fig:training_recipe}
    % \vspace{-0.3cm}
\end{figure*}

\subsection{VLMs for 3D Spatial Intelligence}

Early Vision Language Models achieved remarkable success in 2D visual-language understanding but exhibited clear limitations in spatial reasoning. Recent research has begun addressing this gap from two complementary directions: explicit geometric modeling and 3D perception enhancement. SpatialVLM~\cite{chenSpatialVLMEndowingVisionLanguage2024} was among the first to propose an internet-scale 3D spatial question-answering framework, enabling VLMs to acquire quantifiable spatial reasoning abilities. Building upon this, SpatialRGPT~\cite{chengSpatialRGPTGroundedSpatial2024} and SpatialBot~\cite{caiSpatialBotPreciseSpatial2024} injected depth and region-level 3D features into vision encoders, significantly improving the accuracy of relational judgments involving distance and orientation. From the architectural perspective, SpatialReasoner~\cite{maSpatialReasonerExplicitGeneralizable2025} and SpatialLLM~\cite{maSpatialLLMCompound3DInformed2025} introduced explicit 3D representations and reasoning interfaces, enhancing the interpretability and traceability of reasoning chains. SSR~\cite{liuSSREnhancingDepth2025} further advanced this direction by transforming raw depth information into language-based intermediate ``reasoning texts", thus bridging low-level perception and high-level semantic inference.

More recent efforts, including MM-Spatial~\cite{daxbergerMMSpatialExploring3D2025}, VLM-3R~\cite{fanVLM3RVisionLanguageModels2025}, and Video-3D LLM~\cite{zhengVideo3DLLMLearning2025}, integrated 3D reconstruction with video modeling to unify the representation spaces of frames, point clouds, and text. These approaches established an essential foundation for spatio-temporal 3D understanding. However, most of them still rely on offline reconstruction or fixed 3D inputs, making continuous and interpretable spatial reasoning within dynamic video scenes difficult to achieve.
Our proposed GEODE addresses this limitation by jointly encoding point clouds and video sequences and incorporating a 3D-CoT reasoning mechanism that supports consistent, geometry-aware inference over time.

\subsection{Vision Language Models and Video Understanding}

The rapid development of Vision-Language Models has significantly advanced multimodal reasoning and perception. Spatial-MLLM~\cite{wuSpatialMLLMBoostingMLLM2025} and Multi-SpatialMLLM~\cite{xuMultiSpatialMLLMMultiFrameSpatial2025} introduced geometric priors and multi-frame fusion strategies, granting VLMs preliminary spatio-temporal reasoning capabilities. MindJourney~\cite{yangMindJourneyTestTimeScaling2025} extended this concept by employing world model during inference to generate multi-view evidence, thereby enhancing spatial consistency at test time. Meanwhile, several studies have focused on benchmarking and reinforcing spatial reasoning. OmniSpatial~\cite{jiaOmniSpatialComprehensiveSpatial2025}, Mind-the-Gap~\cite{stogiannidisMindGapBenchmarking2025}, and SpatialLadder~\cite{liSpatialLadderProgressiveTraining2025} constructed hierarchical benchmarks grounded in cognitive and task complexity perspectives, revealing the weaknesses of current models in complex spatial logic tasks. However, GEODE introduces a novel dual-decoupling architecture that resolves the fundamental input-reasoning and output-generation bottlenecks for spatio-temporal reasoning in videos.

% For video-based reasoning, SpaceR~\cite{ouyangSpaceRReinforcingMLLMs} and MetaSpatial~\cite{panMetaSpatialReinforcing3D2025} leveraged reinforcement learning and physical constraints to capture consistent 3D relations across temporal sequences, leading to measurable improvements in interpretable spatio-temporal reasoning. Despite these advances, most existing Video-LLMs remain confined to 2D semantic aggregation and lack explicit alignment with real-world geometry. In contrast, GEODE introduces a novel dual-decoupling architecture that resolves the fundamental input-reasoning and output-generation bottlenecks for spatio-temporal reasoning in videos.
\section{Limitations and Future Work}
\label{sec:limitations}

While our \textbf{GEODE} framework demonstrates a significant leap in parameter-efficient spatial reasoning, we identify several limitations and promising directions for future research.

\noindent\textbf{Limitations.}
Our current framework relies on pretrained, external models for 3D reconstruction VGGT~\cite{wangVGGTVisualGeometry2025} and point cloud encoder Sonata~\cite{wuSonataSelfSupervisedLearning2025}. The overall performance of the DRM is therefore inherently capped by the quality and accuracy of these upstream modules. Furthermore, our ``Embedding-as-Value" paradigm currently requires pre-defined control tokens ($\langle\text{REG}\rangle$, $\langle\text{3DBBOX}\rangle$) for specific tasks. Extending the DRH to novel, unseen regression types without retraining remains an open challenge.

\noindent\textbf{Future Work.}
The GEODE architecture opens several exciting avenues for research. The remarkable parameter-efficiency of our 1.5B model makes it a prime candidate for deployment in resource-constrained environments, providing a robust foundation for \textbf{Embodied AI and Robotics}. The ability to understand 3D space (via DRM) and output precise coordinates (via DRH) is critical for tasks like navigation, manipulation, and human-robot interaction.

Moreover, the ``Embedding-as-Value" concept is highly extensible. We plan to expand the DRH's capabilities beyond scalars and 3D BBoxes to regress more complex and high-dimensional geometric structures. Future work could explore decoding specialized token embeddings into 3D semantic segmentation voxels, dense depth maps, or even continuous navigational trajectories, further bridging the gap between language-based reasoning and continuous-world interaction.
\section{Conclusion}
\label{sec:conc}

In this work, we proposed \textbf{GEODE} which designed to resolve the fundamental bottlenecks VLMs face in 3D spatial intelligence: input-stage reasoning and output-stage numerical generation. Our framework introduces a co-processor to distill 3D spatial logic from CoT data into injectable Rationale Tokens, providing rich, pre-computed context. Simultaneously, our ``Embedding-as-Value" paradigm bypasses tokenization to directly regress precise numerical values and 3D coordinates. This synergistic approach enables parameter-efficient 1.5B model to achieve state-of-the-art spatial reasoning rivaling 7B+ models, which provides robust and efficient foundation for future applicationsmwhich demand precise and grounded spatial perception.
{
    \small
    \bibliographystyle{ieeenat_fullname}
    \bibliography{main}
}


\end{document}